\documentclass[journal]{IEEEtran}
\usepackage{amsmath,amsfonts}
\usepackage{algorithmic}
\usepackage{algorithm}[t]
\usepackage{svg}
\usepackage{multirow}
\usepackage[caption=false,font=normalsize,labelfont=sf,textfont=sf]{subfig}
\begin{document}

\title{Learning Domain Invariant Prompt for Vision-Language Models}

\author{Cairong Zhao, Yubin Wang, Xinyang Jiang, Yifei Shen, Kaitao Song, Dongsheng Li, and Duoqian Miao
\thanks{This work was supported by National Natural Science Fund of China (62076184, 61976158, 61976160, 62076182, 62276190), in part by Fundamental Research Funds for the Central Universities and State Key Laboratory of Integrated Services Networks (Xidian University), in part by Shanghai Innovation Action Project of Science and Technology (20511100700) and Shanghai Natural Science Foundation (22ZR1466700). Cairong Zhao, Yubin Wang contributed equally to this work. Corresponding author is Cairong Zhao.}
\thanks{Cairong Zhao, Yubin Wang and Duoqian Miao are with the Department of Computer Science and Technology, Tongji University, Shanghai 201804, China (e-mail: zhaocairong@tongji.edu.cn; wangyubin2018@tongji.edu.cn; dqmiao@tongji.edu.cn).}
\thanks{Xinyang Jiang, Yifei Shen, Kaitao Song and Dongsheng Li are with Microsoft Research Asia, Shanghai 200232, China (e-mail: xinyangjiang@-microsoft.com; yifeishen@microsoft.com; kaitaosong@microsoft.com; dongsheng.li@microsoft.com).}}

\markboth{Journal of \LaTeX\ Class Files,~Vol.~14, No.~8, August~2021}%
{Shell \MakeLowercase{\textit{et al.}}: A Sample Article Using IEEEtran.cls for IEEE Journals}


\maketitle

\begin{abstract}
   Prompt learning is one of the most effective and trending ways to adapt powerful vision-language foundation models like CLIP to downstream datasets by tuning learnable prompt vectors with very few samples. 
   However, although prompt learning achieves excellent performance over in-domain data, it still faces the major challenge of generalizing to unseen classes and domains. 
   Some existing prompt learning methods tackle this issue by adaptively generating different prompts for different tokens or domains but neglecting the ability of learned prompts to generalize to unseen domains. 
   In this paper, we propose a novel prompt learning paradigm that directly generates \emph{domain invariant} prompt that can be generalized to unseen domains, called MetaPrompt. 
   Specifically, a dual-modality prompt tuning network is proposed to generate prompts for input from both image and text modalities.
   With a novel asymmetric contrastive loss, the representation from the original pre-trained vision-language model acts as supervision to enhance the generalization ability of the learned prompt.
   More importantly, we propose a meta-learning-based prompt tuning algorithm that explicitly constrains the task-specific prompt tuned for one domain or class to also achieve good performance in another domain or class.
   Extensive experiments on 11 datasets for base-to-new generalization and 4 datasets for domain generalization demonstrate that our method consistently and significantly outperforms existing methods.
\end{abstract}


\section{Introduction}
\IEEEPARstart{R}{ecent} research in pre-training large Vision-Language Models (VLM) using web-scale data has shown remarkable progress in learning transferable representations~\cite{radford2021learning, jia2021scaling}. 
Compared with traditional supervised learning methods, which learn close-set visual concepts from discrete labels, these models align images in a joint embedding space via contrastive learning, providing a promising opportunity to leverage human language for guiding visual recognition tasks.
Benefiting from this paradigm, pre-trained vision-language models can conduct zero-shot or few-shot transfer to downstream tasks with open-set visual concepts learned from natural language supervision.
As a result, how to effectively leverage these powerful foundation models becomes an important research direction. 
Recent studies~\cite{zhou2022learning, zhou2022conditional} apply a simple yet effective way to adapt pre-trained vision-language models to downstream tasks, called prompting. 
Manually designing a proper prompt is a non-trivial task due to its ambiguity, which makes automatic prompt tuning the current mainstream approach.
Drawing inspiration from recent advances of prompt learning \cite{li2021prefix, lester2021power, liu2021p} in NLP, methods like CoOp \cite{zhou2022learning}, CoCoOp \cite{zhou2022conditional} and MaPLe \cite{khattak2022maple} learn a set of continuous vectors as the context in a prompt (i.e., prompt vector) with the pre-trained parameters fixed, which achieve significant improvement with very few training samples.

Although showing promising performance in i.i.d samples, as discussed by previous works \cite{zhou2022conditional}, prompt learning still faces a substantial challenge of domain generalization. 
Like other machine learning methods, conventional prompt tuning approaches \cite{zhou2022learning} tend to overfit the distribution of the training set. 
When transferred to unseen domains, the good generalization ability of foundation models is compromised, and the learned prompt vectors suffer a significant accuracy drop when transferred to unseen domains. 
Even with massive tuning, we could not yet guarantee an optimal prompt for the downstream tasks.
Recently, several methods \cite{zhou2022conditional, zheng2022prompt} have tackled this issue by adaptively generating different prompts for different tokens or domains, known as conditional prompt learning.
However, they fail to exploit the generalization ability of the prompt generator or learned prompt, and do not explicitly enforce the prompt to generalize to unseen domains. 

In this paper, our goal is to explicitly learn domain invariant prompt for vision-language models. 
Such prompts are independent of the input instance and should have a low bias toward the visual representations of the target task.
In essence, with the distribution shift in text and image modality, both two can be categorized as cross-domain tasks, as the test samples are out-of-domain. 
As discussed in previous literature \cite{wiles2021fine,higgins2017beta,kim2018disentangling}, input samples are composed of attributes (i.e., factors of variation), such as color, shape, texture, etc., and different domains are defined by different distributions of each attribute. 
As a result, there exists a unified meta-domain containing all possible attributes, where data domains are attribute distributions sampled from this meta-domain. 
Under this assumption, our theoretical analysis following \cite{chen2020closer} shows that tuning prompts with an episodic training strategy has a strong generalization guarantee. 
Specifically, this type of method has the generalization bound of $O(1 / \sqrt{N})$ with $N$ being the number of tasks, independent of the sample size in each domain, which motivates us to propose an episodic prompt tuning method in the few-shot setting.

As a result, leveraging the power of episodic training, we propose MetaPrompt, a simple but effective few-shot approach that generates domain invariant prompt for vision-language models.
To explicitly enforce the learned prompt to generalize to unseen domains, a novel episodic prompt tuning algorithm is proposed, where we optimize the prompt when trained on a certain domain that can produce good results on samples out of this domain. 
Furthermore, a dual-modality prompt tuning network is proposed, where two sets of prompt vectors are learned for input from both image and text modalities, respectively. 
To overcome overfitting on in-domain data and leverage the strong generalization ability of the pre-trained vision-language model, we propose an asymmetric contrastive loss where the representation from the unprompted stream of text modality acts as supervision to learn visual prompt, and vice versa, aiming to learn a set of domain invariant prompts for both modalities. 

\begin{figure}[t]
  \centering
    \center{\includegraphics[width=8.5cm]
    {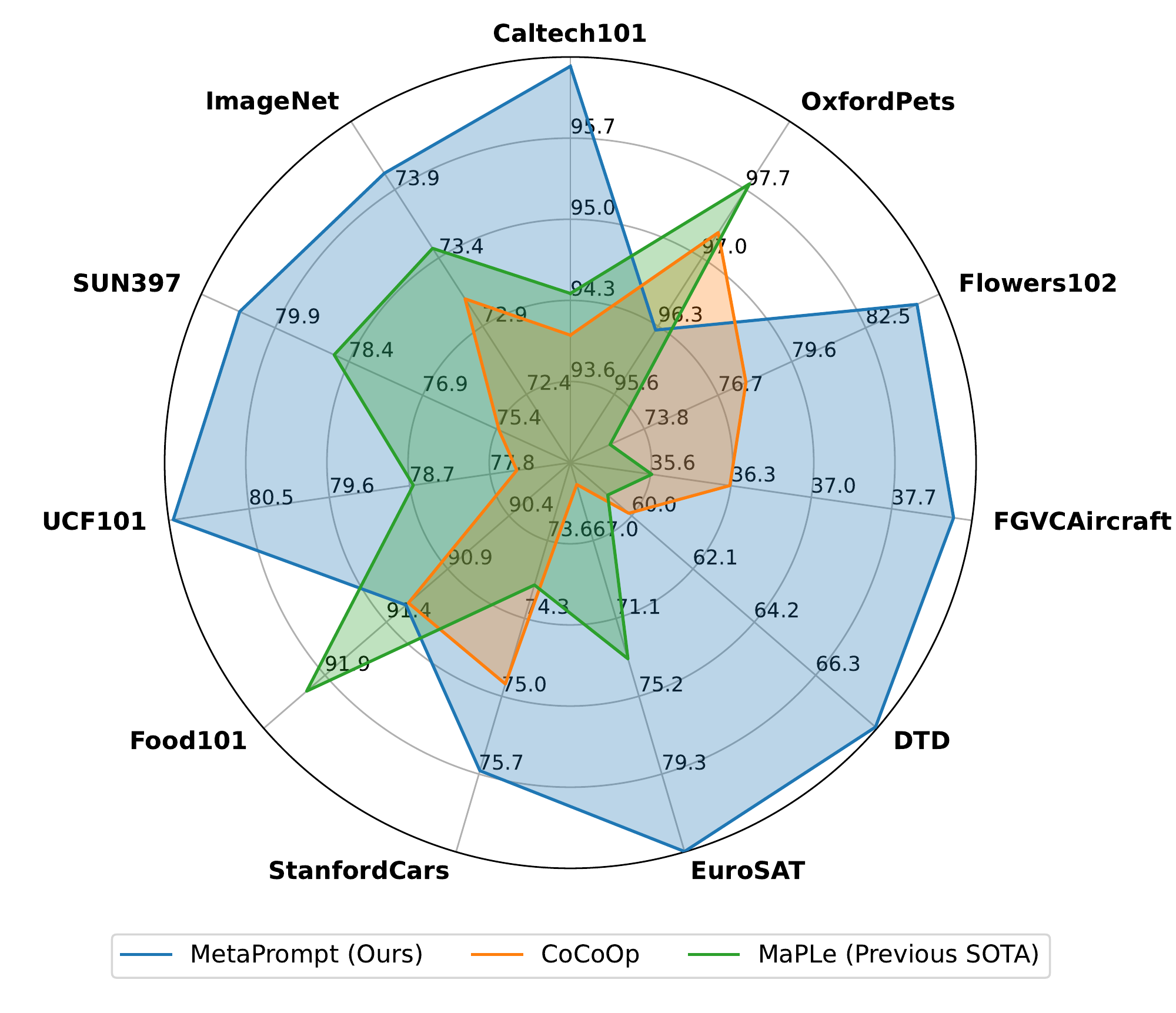}}
    \caption{Comprehensive comparison of the harmonic mean of previous methods CoCoOp, MaPLe, and our method MetaPrompt on 11 diverse image recognition datasets for base-to-new generalization. MetaPrompt surpasses state-of-the-art methods on 9 of 11 datasets, proving that learning domain invariant prompt achieves a good trade-off between in-domain and out-of-domain data.\label{fig:radar}}
\end{figure}

In this paper, the ability of domain generalization is evaluated from two perspectives, new image domains and new class domains. Our MetaPrompt is applicable for both out-of-domain images (i.e., conventional domain generalization task) and classes (i.e., base-to-new generalization task). 
As shown in Fig. \ref{fig:radar}, for base-to-new generalization, MetaPrompt obtains an overall improvement of harmonic mean accuracy by an average gain of 0.54\% over the previous state-of-the-art method MaPLe on 11 image recognition datasets. 
For domain generalization, although based on a few-shot setting, our method achieves comparable performance over other methods training on full datasets with only 1-shot and 5-shot settings.
These experimental results demonstrate the effectiveness of our method and show superiority in generalization ability to other prompt tuning approaches.

\section{Related Work}
\subsection{Prompt Learning} Prompt learning emerges from recent advances in natural language processing. 
The core idea of prompt learning is to formalize various tasks \cite{devlin2018bert, radford2021learning, radford2019language} to masked language modeling problems with different prompt templates. 
A prompt can be seen as a function of the input tokens, providing instruction for pre-trained language models such as BERT \cite{devlin2018bert} or GPT \cite{radford2019language} to adapt to downstream tasks. 
Earlier work \cite{liu2023pre} enabled the model to understand the task and make better predictions by manually designing discrete natural language prompts. 
However, some hand-crafted prompt templates are inappropriate in many cases due to their ambiguity, and the recognition performance of such methods is susceptible to the form of the provided content. 
Recent methods~\cite{li2021prefix, lester2021power, liu2021p} learn continuous contexts to model prompts, called prompt tuning, to automate prompt engineering and explore optimal prompts. This paradigm can also be applied to vision-language models \cite{radford2021learning, jia2021scaling}. Specifically, CoOp~\cite{zhou2022learning} demonstrates that the performance of CLIP is sensitive to prompts, and a suitable prompt can be learned with very few samples for image recognition. CoCoOp \cite{zhou2022conditional} extends CoOp by learning an input-conditional token for each image to obtain generalizable representations. ProDA \cite{lu2022prompt} captures the distribution of diverse prompts to handle the varying visual representations and provides high-quality task-related content for facilitating recognition.

Although these approaches consider prompt learning for text modality, they neglect to tune prompts for generating visual features. To fill this gap, Visual Prompt Tuning (VPT)~\cite{jia2022visual} achieves significant performance gains with only a small amount of trainable parameters as a prompt while keeping the model backbone frozen. MaPLe \cite{khattak2022maple} proposes a prompt tuning method for both vision and language branches to improve alignment between the vision and language representations.
In contrast, with an explicit constraint on prompt tuning, our method learns domain invariant prompt for both modalities, resulting in better generalization.

\subsection{Domain Generalization} 
Domain generalization refers to learning a robust model generalized to unseen domains. In this paper, the generalization ability of a model is evaluated from the perspectives of both out-of-domain images and classes, corresponding to the conventional domain generalization task and base-to-new generalization task. Conventional domain generalization mainly evaluates the generalization ability on unseen image domains. 
Many approaches~\cite{li2018domain, ganin2015unsupervised, motiian2017unified, arjovsky2019invariant} attempt to measure the domain gap and learn domain invariant features.
In order to learn a set of parameters that can generalize to unseen domains, several methods~\cite{balaji2018metareg,li2018learning} adopt meta-learning to simulate domain shift during training. 
In this paper, we provide a theoretical analysis on the generalization guarantee of meta-learning based on episodic training and incorporate episodic training in prompt tuning for the first time.
In contrast to previous methods, we treat the regularization process for generalization as a constraint after regular steps separately, aiming to create various episodes within one single batch to optimize the domain invariant prompt with less computational complexity.

Recently, another type of generalization task emerges called base-to-new generalization, which aims to exploit the generalization ability on unseen classes \cite{chao2016empirical, wang2019survey, xian2017zero, xian2018feature}.
Conventional methods \cite{huynh2020fine, frome2013devise, wang2018zero, kampffmeyer2019rethinking} learn a semantic space based on auxiliary information.
Compared with supervised learning, CLIP-based methods achieve high performance in generalization due to the more vital transferring ability. CoCoOp \cite{zhou2022conditional} attempts to tackle this generalization problem with conditional prompt learning. 
We investigate the feasibility of learning domain invariant prompts for the pre-trained vision-language model CLIP \cite{radford2021learning} and propose a training strategy to implement this goal.
\subsection{Meta-Learning} Most existing meta-learning approaches focus on few-shot learning, which can be divided into metric learning methods, memory network methods, and optimization-based methods. 
Metric learning methods \cite{vinyals2016matching, snell2017prototypical, sung2018learning, rodriguez2020embedding} learn a similarity space to extract discriminative meta-features for new classes efficiently. 
Memory network methods \cite{mishra2017simple, munkhdalai2017meta, oreshkin2018tadam, santoro2016meta} store meta-knowledge by memory models when learning seen tasks and then generalize it to unseen tasks. 
Optimization-based methods \cite{finn2017model, ravi2017optimization, finn2018probabilistic, rajeswaran2019meta} train meta-optimizer that enable fast adaption for new tasks. 
Works like MAML \cite{finn2017model,antoniou2018train, flennerhag2019meta, zhou2019efficient}  focus on learning meta-initial parameters of a deep model so that it would perform well on new tasks after only a small number of gradient updates. 
Drawing on the advances, our work is the first to propose learning meta-prompt for visual-language foundation models generalizing to unseen domains. Instead of learning the initial parameters of the model, we regularize parameters after every conventional update to learn robust representations. In concrete, we utilize gradients on meta-test subtasks to regularize parameters, i.e., prompts. By imposing this constraint, our model learns robust representations and performs better on base-to-new generalization and domain generalization tasks.

\begin{figure}[t]
  \centering
    \center{\includegraphics[width=8cm]
    {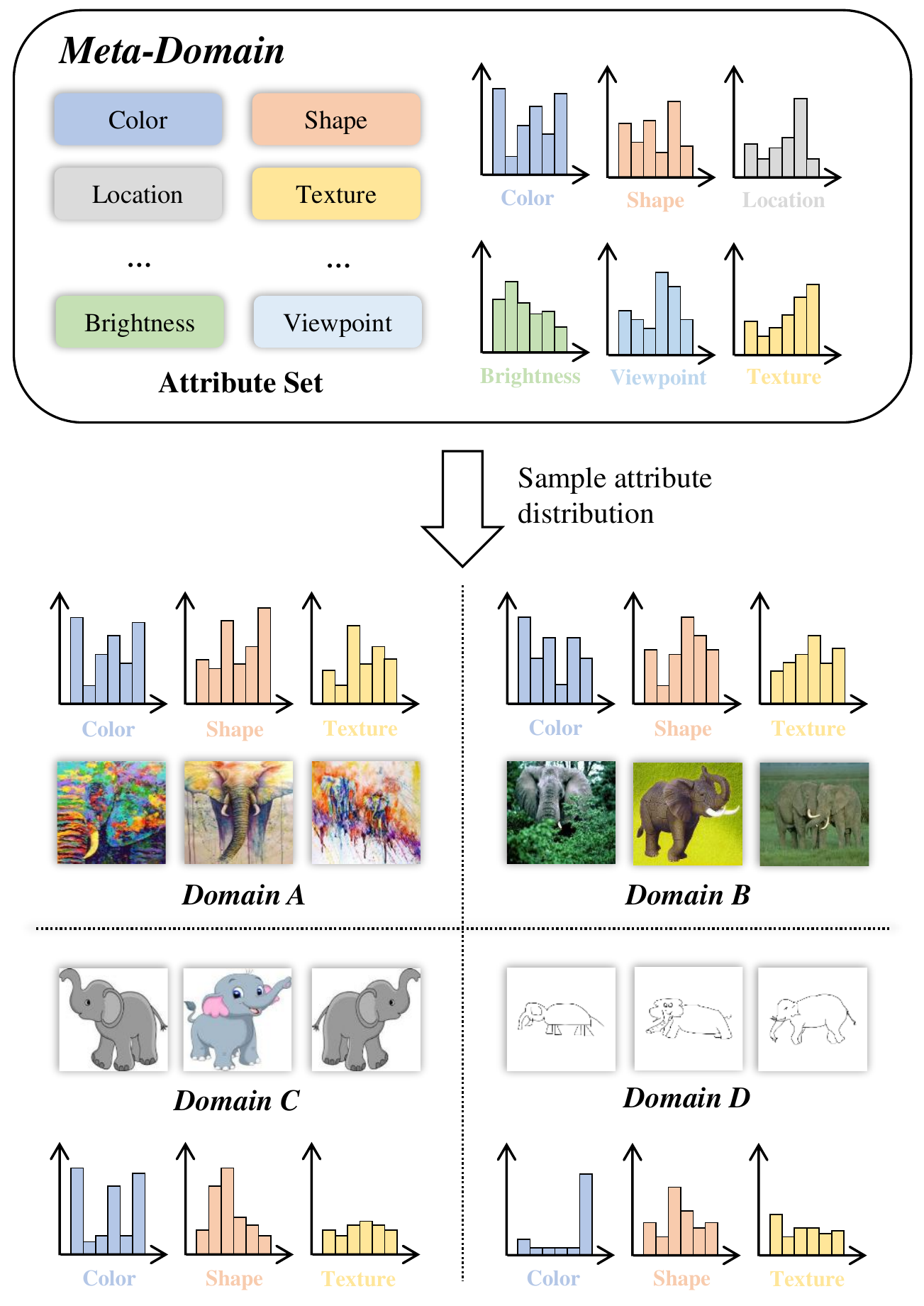}}
    \caption{Input samples are composed of attributes (i.e., factors of variation), such as color, shape, texture, etc., and different domains can be defined by different distributions of attributes.\label{fig:meta_domain}}
\end{figure}

\begin{figure*}[!t]
    \centering
    \center{\includegraphics[width=13cm]{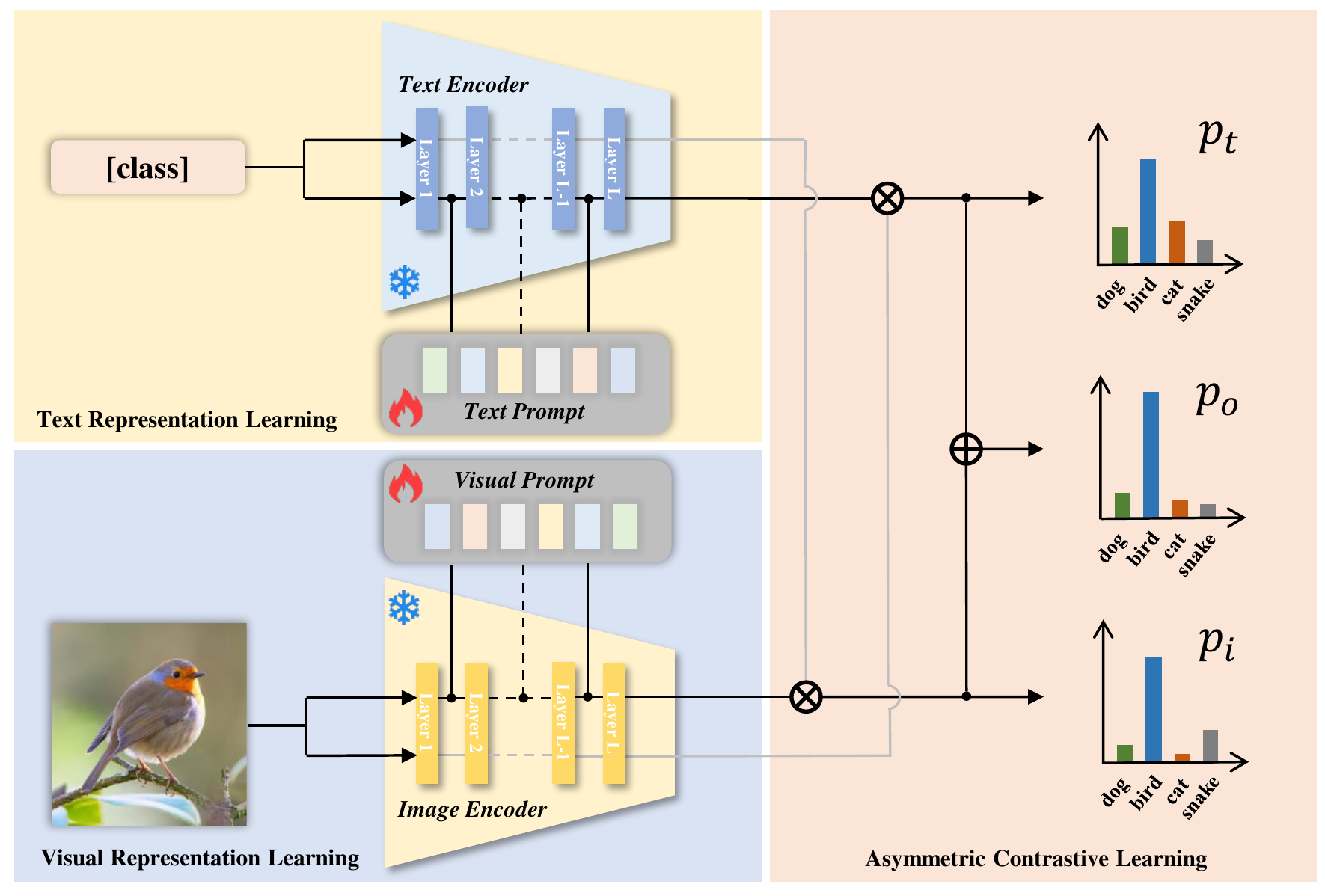}}
    \caption{Our dual-modality prompt tuning network consists of a text encoder for textual representation learning and an image encoder for visual representation learning. Each encoder has two streams, where the unprompted stream (marked by the gray line) from one modality guides the prompted stream (marked by the black line) from another modality by tuning corresponding prompts. The asymmetric contrastive learning module outputs three probability distributions for the end-to-end training to make better predictions.\label{fig:model}}
\end{figure*} 

\section{Generalization Bound of Episodic Training}
Following previous literature~\cite{wiles2021fine}, our theoretical analysis is based on the assumption that data is composed of attributes (i.e., factors of variation), such as color, shape, texture, etc., and different domains can be defined by different distributions of attributes. 
For example, as shown in Fig. \ref{fig:meta_domain}, a sketch domain corresponds to a color distribution with only two values, black and white. In contrast, a cartoon or natural image domain may correspond to a color distribution with more color values. 
As a result, we assume that there exists a unified meta-domain distribution $\tau$ containing all possible attributes, where data domains $\mathcal{P}=\{\mathcal{P}_i\}_{i=1}^N$ are distributions sampled from this meta-domain with different attribute distributions. 
Under this assumption, we expect a training strategy to learn invariant features from seen domains and be able to generalize to unseen domains. Specifically, given a training algorithm $\mathbf{F}$ trained on a dataset $\mathbf{D}=\left\{D_i=D^{s}_i\right\}_{i=1}^N$  drawn from a domain distribution $\mathcal{P}^M_i$ containing $M$ training samples (i.e., $D_i^s \stackrel{i . i . d .}{\sim} \mathcal{P}^M_i$), 
 the generalization error of the model obtained by  $\mathbf{F}(\mathbf{D})$ is as follows:
 
 \begin{equation}
 \mathcal{R}(\mathbf{F}(\mathbf{D}), \tau)=\mathbb{E}_{\mathcal{P}\sim\tau,D^{s} \sim \mathcal{P}^M,z\sim\mathcal{P}}L(\mathbf{F}(\mathbf{D})(D^{s}), z). 
 \end{equation}

To improve the generalization ability of meta-learning algorithms, the pioneering work \cite{vinyals2016matching} proposes a training strategy – episodic training strategy, which treats each task as a training instance and updates the inner-task algorithm by episode (task by task). 
In this paper, we transfer episodic training to the domain generalization scenario by treating each data domain as a training instance and update the inner-domain algorithm by episode (domain by domain). Specifically, we first update the model on a support domain (i.e., in-domain error). Then the performance of the updated model is measured and optimized on another query domain (i.e., out-of-domain error or episodic training error). 
As a result, the training error of the episodic training strategy is as follows: 

\begin{equation}
\hat{\mathcal{R}}_{epi}(\mathbf{F}(\mathbf{D}), \mathbf{D}) = \frac{1}{N}\sum_{i=1}^N\frac{1}{N_i^{q}}\sum_{z_{i}\in D_{i}^{q}}\hat{L}(\mathbf{F}(D_i^{s}), z_{i}), 
\label{eq_ep_error}
\end{equation}
where $D_{i}^{q}$ is the set of data sampled from a query domain; and $N_{i}^{q}$ is the number of samples in $D^{q}_i$.  
From Eq.~\ref{eq_ep_error} we can see that episodic training strategy directly minimizes the out-of-domain testing error, and hence intuitively the in-domain sample number $M$ in the generalization bound vanishes, with the generalization bound only depending on the domain number $N$.  

Based on this paradigm, we naturally associate episodic training with domain generalization task, aiming to learn invariance from various distributions by creating meta-tasks with domain gap as episodes.
By applying this strategy, the distribution shift between the meta-train and meta-test subtask can be approximately equivalent to that between the original training and test task. 
The error of the parameter over the meta-test task is exactly the test error of generalization tasks and thereby is an unbiased estimate of the generalization error on unseen domains.        
Theoretically, following \cite{chen2020closer}, we derive the bound of the generalization gap between these two errors only depending on the domain number $N$, which is formulated by:
\begin{equation}
    \mathbb{E}_{\mathbf{F}}[\mathcal{R}(\mathbf{F}(\mathbf{D}), \tau)] \leq \mathbb{E}_{\mathbf{F}}\left[\hat{\mathcal{R}}_{epi}(\mathbf{F}(\mathbf{D}), \mathbf{D})\right]+O\left({\frac{1}{\sqrt{N}}}\right).
\end{equation}

The generalization bound implies a strong generalization guarantee for episodic training algorithms in the few-shot regime, which motivates this paper to adopt episodic training to learn domain invariant prompt with very few samples.

\section{Methodology}
In this section, we elaborate on our prompt tuning method, MetaPrompt. 
Sec.~4.1 provides a brief overview of existing prompt tuning approaches for text and image modalities. 
Sec.~4.2 introduces our dual-modalities prompt tuning network and demonstrates a novel asymmetric contrastive loss for prompt tuning.
Sec.~4.3 introduces our batch-wise episodic training paradigm for learning invariant prompt.
\subsection{Dual-Modality Prompt Tuning}
This paper focuses on prompt tuning for vision-language foundation models, such as CLIP, which are usually composed of a text encoder and an image encoder. 
The text encoder adopts a transformer \cite{vaswani2017attention} to encode textual information while the image encoder can either be a CNN model like ResNet \cite{he2016deep} or a vision transformer like ViT \cite{dosovitskiy2020image} to encode visual concepts. 
Among recent works on prompt tuning, prompt vectors can be learned for both text \cite{zhou2022learning} and image encoder \cite{jia2022visual}. We describe the setting of prompt tuning for text and image modalities as follows:
\paragraph{Textual Prompt Tuning}\cite{zhou2022learning} automatically learns a set of tunable continuous vectors as context tokens that are fed into the text encoder together with the class tokens.
Given the textual prompt composed of $P$ vectors for the $i$-th class denoted as $\boldsymbol{t}_i$, the prediction probability of the $i$-th class can be calculated by:
\begin{equation}
\label{eq2}
    p_t(y=i \mid \boldsymbol{x})=\frac{\exp \left(\operatorname{sim}\left(\boldsymbol{x}, g\left(\boldsymbol{t}_i\right)\right) / \tau\right)}{\sum_{j=1}^K \exp \left(\operatorname{sim}\left(\boldsymbol{x}, g\left(\boldsymbol{t}_j\right)\right) / \tau\right)},
\end{equation}
where $\boldsymbol{x}$ represents the image feature extracted from the image encoder and $g(\cdot)$ denotes the text encoder.
\paragraph{Visual Prompt Tuning}\cite{jia2022visual} adopts a similar idea as textual prompt, where extra prompt vectors are automatically learned to be fed into the image encoder.
The image patches are firstly embedded into a latent space as the input of the first transformer layer, and then $P$ learnable vectors are introduced at $L$ transformer layers’ input space as prompt. 
The output of the transformer head is considered the final visual feature $\widetilde{\boldsymbol{x}}$.
The prediction probability of the $i$-th class can be calculated by:
\begin{equation}
\label{eq3}
p_i(y=i \mid \boldsymbol{x})=\frac{\exp \left(\operatorname{sim}\left(\widetilde{\boldsymbol{x}}, g\left(\boldsymbol{h}_i\right)\right) / \tau\right)}{\sum_{j=1}^K \exp \left(\operatorname{sim}\left(\widetilde{\boldsymbol{x}}, g\left(\boldsymbol{h}_j\right)\right) / \tau\right)},
\end{equation}
\noindent where $g(\cdot)$ denotes the text encoder and $\boldsymbol{h}_i$ denotes the textual feature of the $i$-th class token. 

\subsection{Asymmetric Contrastive Learning}
Motivated by previous works on textual and visual prompt tuning, we propose a dual-modality prompt tuning network that jointly learns visual and textual prompts for each transformer layer. 
As shown in Fig. \ref{fig:model}, two sets of learned prompt vectors are fed into the text and image encoder of a foundation model together with the image and text input, respectively.

To prevent the learned prompt vectors from overfitting the in-domain training samples (especially in a few-shot learning setting), we propose to leverage the strong generalization ability of the pre-trained vision language model by a novel Asymmetric Contrastive Loss (AC Loss).
AC loss utilizes representations from the unprompted pre-trained vision-language model as supervision to enhance the generalization ability of prompts and learn a set of domain invariant prompts for both modalities. 
Specifically, instead of training both textual and visual prompts simultaneously with a single contrastive loss, we train them separately, where prompted representations from one modality are aligned with unprompted ones from another modality. 
For example, as shown in Fig. \ref{fig:model}, for learning of visual prompt, the prediction probability is obtained based on the similarity between unprompted textual features and prompted visual features. 

With our asymmetric contrastive learning module, we have two probability distributions $p_t$ and $p_i$, corresponding to textual and visual prompts with Eq. \ref{eq2} and Eq. \ref{eq3}. 
We sum them up to obtain an overall probability distribution $p_o$, which is formulated by:

\begin{equation}
p_o(y \mid \boldsymbol{x})=\frac{1}{2}(p_t(y \mid \boldsymbol{x})+p_i(y \mid \boldsymbol{x})).
\end{equation}

During training, the cross-entropy loss is adopted to minimize the distance between the ground-truth label $y$ and the three probability distributions $p_t$, $p_i$ and $p_o$, where the losses are denoted as $\mathcal{L}_t$, $\mathcal{L}_i$ and $\mathcal{L}_o$. As a result, AC Loss can be expressed as the sum of the above three losses:
 \begin{equation}
     \mathcal{L}_{AC}(y, \boldsymbol{x}) = \mathcal{L}_{o}(y, \boldsymbol{x})+\mathcal{L}_{t}(y, \boldsymbol{x})+\mathcal{L}_{i}(y, \boldsymbol{x})
 \end{equation}
 
\begin{figure}[t]
    \centering
    \center{\includegraphics[width=8cm]{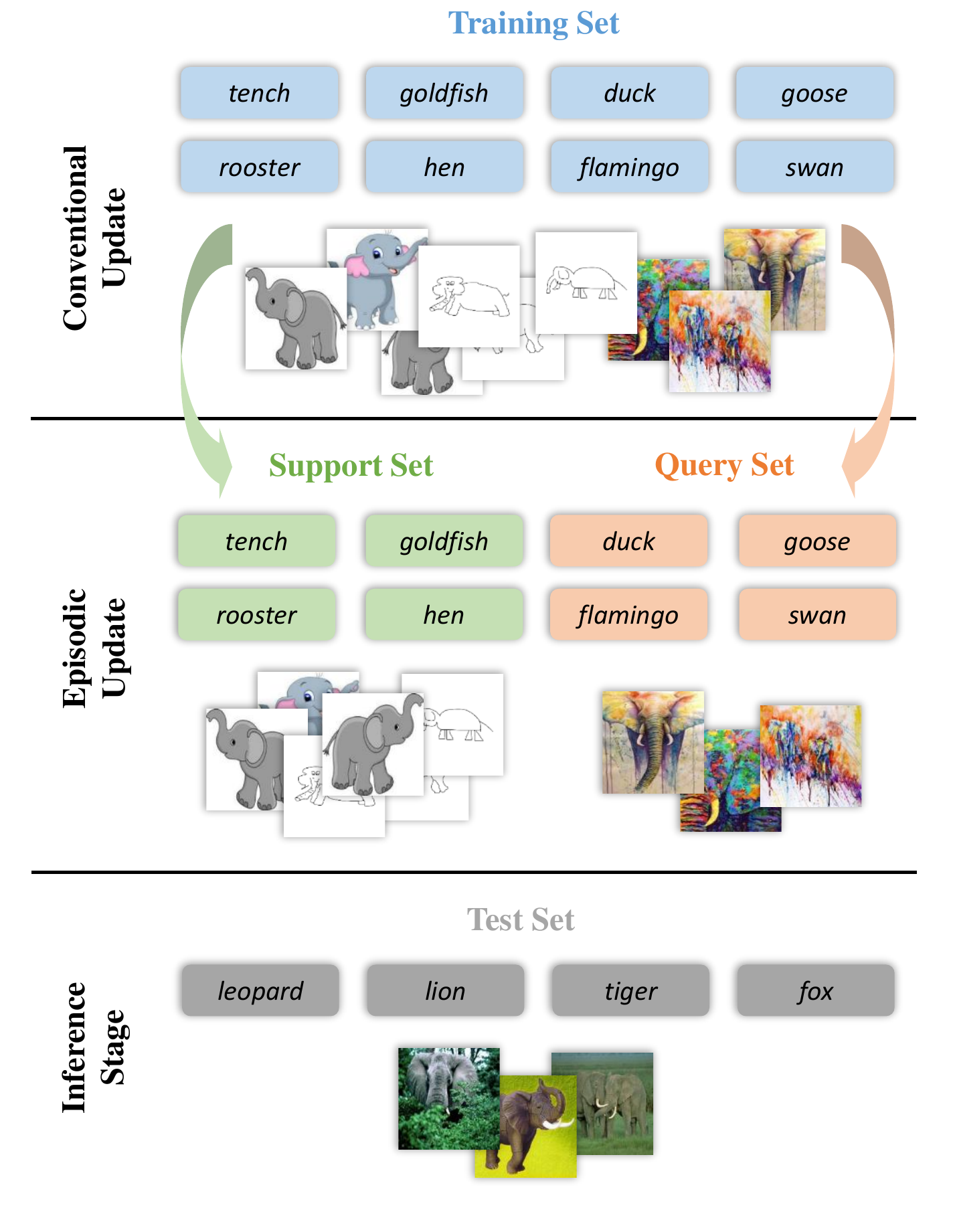}}
    \caption{Given a batch of training data containing samples from different domains, we conduct a conventional update and several episodic updates. In the inference stage, with the domain invariant prompt, we evaluate the generalization ability on unseen domains. \label{fig:sep}}
\end{figure} 

\subsection{Batch-wise Episodic Training}
Motivated by the analysis from Section 3, we propose a batch-wise episodic training paradigm for prompt learning.
Given a batch of training data containing samples from different domains, we separate the batch into a support set and a query set based on domains. 
Our proposed episodic training strategy aims to regularize learnable prompts that narrow the gap between training errors on the support set and query set.

Specifically, given a set of $N$ datasets sampled from $N$ domains denoted as $\mathbf{D}=\{D_i\}_{i=1}^N$ where $D_i \sim \mathcal{P}_i$, we split the set by grouping samples from one of the datasets as the query set $D_i^{q}$, and samples from the rest as the support set $D_i^{s}$.
Note that, for domain generalization, since it is clear which domain each sample belongs to, the query and support set can be easily split.
However, for base-to-new generalization, there is no explicit definition of which domain each sample belongs to. Hence, we randomly split the query and support set based on the class label of each sample, as shown in Fig. \ref{fig:sep}. 

\begin{algorithm}[t]
\caption{Batch-wise Episodic Training\label{alg1}}
\begin{algorithmic}[1]
\REQUIRE Domain size $N$, learning rate $\eta$, meta-step size $\alpha$, dataset $\mathcal{D}$, loss functions $\mathcal{L}_{AC}$, $\mathcal{L}_{t}$, $\mathcal{L}_{i}$
\ENSURE Prompt parameters $ \Theta$
\STATE Randomly initialize $\Theta_0 = \{\theta^I, \theta^T\}$\
    \FOR {$t$ in iterations}
    
        \STATE Randomly sample a batch $\mathcal{D}_t$ from $\mathcal{D}$\
 	\STATE \textbf{Conventional Update:}
	\STATE Update $ \Theta_t $ w.r.t. $\mathcal{L}_{AC}$: \\
          \quad $\Theta_t \leftarrow \Theta_t-\eta \nabla_{\Theta_t} \mathcal{L}_{AC}\left(\Theta_t;\mathcal{D}_t\right)$\
        \STATE \textbf{Episodic Update:}
        \IF {base-to-new generalization}
		\STATE $\theta \leftarrow \theta_t^T$, $\mathcal{L} \leftarrow \mathcal{L}_t$\
            \STATE $\mathbf{D}=\{D_i\}_{i=1}^N \leftarrow {group\_by\_class(\mathcal{D}_t)}$\
	\ELSIF {conventional domain generalization}
    	\STATE $\theta \leftarrow \theta_t^I$, $\mathcal{L} \leftarrow \mathcal{L}_i$\
            \STATE $\mathbf{D}=\{D_i\}_{i=1}^N \leftarrow {group\_by\_domain(\mathcal{D}_t)}$\
        \ENDIF

        \FOR {$i=1$ to $N$}
            \STATE $D_i^{q} \leftarrow D_{i}$\
            \STATE $D_i^{s} \leftarrow \bigcup_{j=1,j \ne i}^ND_{j}$\
            \STATE $\theta'_i \leftarrow \theta-\alpha \nabla_{\theta} \mathcal{L}\left(\theta;D_i^{s}\right)$\
            \STATE $g_i \leftarrow \nabla_{\theta'_i} \mathcal{L}\left(\theta'_i;D_i^{q}\right)$
        \ENDFOR
        
        \STATE Update $ \theta $ with gradients: \\
            \quad $\theta \leftarrow \theta-\alpha\eta \sum_{i=1}^{N} g_i$
        \IF {base-to-new generalization}
        \STATE $\Theta_{t+1} \leftarrow \{\theta_t^I, \theta\}$
        \ELSIF {conventional domain generalization}
        \STATE $\Theta_{t+1} \leftarrow \{\theta, \theta_t^T\}$
        \ENDIF

    \ENDFOR
\end{algorithmic}
\end{algorithm}

To maintain good in-domain performance, which is an essential metric for base-to-new generalization, we adopt an alternated update strategy, where the query/support episodic update and the conventional gradient descent update are conducted alternately.
After a conventional update, the learnable prompt $\theta$ is updated with the samples on the support set $D_i^{s}$ to get the updated prompt $\theta'_i$.
Then the generalization error of the updated prompt $\theta'_i$ is measured by the cross-entropy loss on the query set $D_i^{q}$, whose corresponding gradients are back-propagated to update the original prompt $\theta$.
Since this update involves second-order gradient computation with high complexity, in our implementation, we design a first-order approximating method. 
The parameter $\theta$ is updated as follows: 

\begin{equation}
\theta \leftarrow \theta-\alpha\eta \sum_i \nabla_{\theta'_i} \mathcal{L}(\theta'_i;D_i^{q}),
\label{eq_gradient}
\end{equation}
where $\alpha$ is the meta-step size, and $\eta$ is the learning rate of the conventional training. $\mathcal{L}$ indicates the generalization loss on the query set for calculating gradients.
To simplify the training process, our paradigm treats one batch-wise iteration in Eq.~\ref{eq_gradient} as a series of training episodes and conducts several splits of the query and support set within each batch iteration.

Furthermore, based on our dual-modality prompt tuning network, we propose a modality-specific optimization strategy where the prompt of only one specific modality is tuned during the episodic update step. 
For example, since base-to-new generalization focuses on generalizing to unseen classes, only textual prompts are tuned with the loss function $\mathcal{L}_t$.
Similarly, for conventional domain generalization, only visual prompts are updated with the loss function $\mathcal{L}_i$ to generalize to unseen images. The detailed implementation of the episodic training is shown in Alg. \ref{alg1}. 

\section{Experiments}
We evaluate our approach mainly in the two generalization settings, i.e. base-to-new generalization and conventional domain generalization. 
In our experiments, we use the open-source CLIP \cite{radford2021learning} as the foundation vision-language model. 
Here we elaborate on the experimental configurations. 

\paragraph{Datasets} For base-to-new generalization, we follow Zhou et al. \cite{zhou2022learning} and evaluate the performance of our method using 11 image recognition datasets, which cover a wide range of recognition tasks. 
Specifically, the benchmark includes ImageNet \cite{deng2009imagenet} and Caltech101 \cite{fei2004learning} for classification on generic objects; OxfordPets \cite{parkhi2012cats}, StanfordCars \cite{krause20133d}, Flowers102 \cite{nilsback2008automated}, Food101 \cite{bossard2014food} and FGVCAircraft \cite{maji2013fine} for fine-grained classification; SUN397 \cite{xiao2010sun} for scene recognition; UCF101 \cite{soomro2012ucf101} for action recognition; DTD \cite{cimpoi2014describing} for texture classification; and finally EuroSAT \cite{helber2019eurosat} for satellite imagery recognition. For each dataset, we split the classes equally into two groups as base and new classes. We train the model only on base classes in a few-shot setting, while evaluation is conducted separately on base and new classes. 

For conventional domain generalization experiments, we select four real-world datasets from DomainBed benchmark, including VLCS~\cite{fang2013unbiased}, PACS~\cite{li2017deeper}, OfficeHome \cite{venkateswara2017deep}, DomainNet \cite{peng2019moment}. We conduct experiments with the leave-one-out strategy. For a dataset, one of the domains is selected as the target domain at a time, and other domains are used as the source domains. We train the model on the source domains in a few-shot setting and evaluated on the target domain.

\paragraph{Implementation Details} We apply prompt tuning on the pre-trained CLIP model with ViT-B/16 as the visual backbone.
Both prompts are randomly initialized from Gaussian distribution with mean of 0 and a standard deviation of 0.02. 
We adopt SGD optimization with an initial learning rate of 0.002, decayed by the cosine annealing rule, and the meta-step size $\alpha$ is set to 0.2.
The warming-up trick is adopted during the first epoch with a fixed learning rate of $10^{-5}$.
During inference, we use the overall distribution $p_o$ for prediction. 

For base-to-new generalization, the maximum epoch is set to 10 for all datasets with a batch size of 16. The prompt length $P$ and the prompt layer $L$ of visual and textual prompts are set to 2 and 12.
Following Zhou et al. \cite{zhou2022learning}, we use the few-shot evaluation protocol selecting 16 shots for training and the whole test set for evaluation.
For conventional domain generalization, the maximum epoch is set to 5 for all datasets with a batch size of 32. The prompt length $P$ and the prompt layer $L$ of visual and textual prompts are set to 4 and 10. 
We adopt 1 and 5 shots for each source domain combining them as the training set and test on the target domain.
For the hyper-parameter selection of our implementation, we share the same set of hyper-parameters instead of searching for each dataset. 
\newcommand\x{2}
\newcommand\y{8}
\begin{table*}[!t]
\caption{\textbf{Comparison of CLIP, CoOp, CoCoOp, MaPLe, and our MetaPrompt on base-to-new generalization.} Our experiments are repeated three times using different random seeds. MetaPrompt outperforms all other methods on both base and new classes and demonstrates strong generalization performance on 11 image recognition datasets. H: Harmonic mean (to highlight the generalization trade-off).\label{tab:b2n}}
	\centering
	\begin{minipage}{0.32\textwidth}
		\centering
            \normalsize
            \textbf{(a) Average over 11 datasets.}
            \vspace{\x pt} 
            
		\begin{tabular}{lcc|c}
            \hline & \textbf{Base} & \textbf{New} & \textbf{H} \\
            \hline CLIP & 69.34 & 74.22 & 71.70 \\
            CoOp & 82.69 & 63.22 & 71.66 \\
            CoCoOp & 80.47 & 71.69 & 75.83 \\
            MaPLe & 82.28 & 75.14 & 78.55 \\
            \hline
            MetaPrompt & \textbf{83.65} & \textbf{75.48} & \textbf{79.09} \\
            vs. MaPLe & \textcolor{red}{+1.37} & \textcolor{red}{+0.34} & \textcolor{red}{+0.54} \\
            \hline
            \end{tabular}
            \vspace{\y pt}
	\end{minipage}\quad
        \begin{minipage}{0.32\textwidth}
		\centering
            \normalsize
            (b) ImageNet.
            \vspace{\x pt} 
            
		\begin{tabular}{lcc|c}
            \hline & \textbf{Base} & \textbf{New} & \textbf{H} \\
            \hline CLIP & 72.43 & 68.14 & 70.22 \\
            CoOp & 76.47 & 67.88 & 71.92 \\
            CoCoOp & 75.98 & 70.43 & 73.10 \\
            MaPLe & 76.66 & 70.54 & 73.47 \\
            \hline
            MetaPrompt & \textbf{77.52} & \textbf{70.83} & \textbf{74.02} \\
            vs. MaPLe & \textcolor{red}{+0.86} & \textcolor{red}{+0.29} & \textcolor{red}{+0.55} \\
            \hline
            \end{tabular}
            \vspace{\y pt}
	\end{minipage}\quad
 	\begin{minipage}{0.32\textwidth}
		\centering
            \normalsize
            (c) Caltech101.
            \vspace{\x pt} 
            
		\begin{tabular}{lcc|c}
            \hline & \textbf{Base} & \textbf{New} & \textbf{H} \\
            \hline CLIP & 96.84 & 94.00 & 95.40 \\
            CoOp & 98.00 & 89.81 & 93.73 \\
            CoCoOp & 97.96 & 93.81 & 95.84 \\
            MaPLe & 97.74 & 94.36 & 96.02 \\
            \hline
            MetaPrompt & \textbf{98.13} & \textbf{94.58} & \textbf{96.32} \\
            vs. MaPLe & \textcolor{red}{+0.39} & \textcolor{red}{+0.22} & \textcolor{red}{+0.30} \\
            \hline
            \end{tabular}
            \vspace{\y pt}
	\end{minipage}
 
	\begin{minipage}{0.32\textwidth}
		\centering
            \normalsize
            (d) OxfordPets.
            \vspace{\x pt} 
            
		\begin{tabular}{lcc|c}
            \hline & \textbf{Base} & \textbf{New} & \textbf{H} \\
            \hline CLIP & 91.17 & 97.26 & 94.12 \\
            CoOp & 93.67 & 95.29 & 94.47 \\
            CoCoOp & 95.20 & 97.69 & 96.43 \\
            MaPLe & 95.43 & \textbf{97.76} & \textbf{96.58} \\
            \hline
            MetaPrompt & \textbf{95.53} & 97.00 & 96.26 \\
            vs. MaPLe & \textcolor{red}{+0.10} & \textcolor{blue}{-0.76} & \textcolor{blue}{-0.32} \\
            \hline
            \end{tabular}
            \vspace{\y pt}
	\end{minipage}\quad
        \begin{minipage}{0.32\textwidth}
		\centering
            \normalsize
            (e) StanfordCars.
            \vspace{\x pt} 
            
		\begin{tabular}{lcc|c}
            \hline & \textbf{Base} & \textbf{New} & \textbf{H} \\
            \hline CLIP & 63.37 & 74.89 & 68.65 \\
            CoOp & \textbf{78.12} & 60.40 & 68.13 \\
            CoCoOp & 70.49 & 73.59 & 72.01 \\
            MaPLe & 72.94 & 74.00 & 73.47 \\
            \hline
            MetaPrompt & 76.34 & \textbf{75.01} & \textbf{75.48} \\
            vs. MaPLe & \textcolor{red}{+3.40} & \textcolor{red}{+1.01} & \textcolor{red}{+2.01} \\
            \hline
            \end{tabular}
            \vspace{\y pt}
	\end{minipage}\quad
 	\begin{minipage}{0.32\textwidth}
		\centering
            \normalsize
            (f) Flowers102.
            \vspace{\x pt} 
            
		\begin{tabular}{lcc|c}
            \hline & \textbf{Base} & \textbf{New} & \textbf{H} \\
            \hline CLIP & 72.08 & \textbf{77.80} & 74.83 \\
            CoOp & 97.60 & 59.67 & 74.06 \\
            CoCoOp & 94.87 & 71.75 & 81.71 \\
            MaPLe & 95.92 & 72.46 & 82.56 \\
            \hline
            MetaPrompt & \textbf{97.66} & 74.49 & \textbf{84.52} \\
            vs. MaPLe & \textcolor{red}{+1.74} & \textcolor{red}{+2.03} & \textcolor{red}{+1.96} \\
            \hline
            \end{tabular}
            \vspace{\y pt}
	\end{minipage}

	\begin{minipage}{0.32\textwidth}
		\centering
            \normalsize
            (g) Food101.
            \vspace{\x pt} 
            
		\begin{tabular}{lcc|c}
            \hline & \textbf{Base} & \textbf{New} & \textbf{H} \\
            \hline CLIP &  90.10 & 91.22 & 90.66 \\
            CoOp & 88.33 & 82.26 & 85.19 \\
            CoCoOp & 90.70 & 91.29 & 90.99 \\
            MaPLe & 90.71 & \textbf{92.05} & \textbf{91.38} \\
            \hline
            MetaPrompt & \textbf{90.74} & 91.85 & 91.29 \\
            vs. MaPLe & \textcolor{red}{+0.03} & \textcolor{blue}{-0.20} & \textcolor{blue}{-0.09} \\
            \hline
            \end{tabular}
            \vspace{\y pt}
	\end{minipage}\quad
        \begin{minipage}{0.32\textwidth}
		\centering
            \normalsize
            (h) FGVCAircraft.
            \vspace{\x pt} 
            
		\begin{tabular}{lcc|c}
            \hline & \textbf{Base} & \textbf{New} & \textbf{H} \\
            \hline CLIP & 27.19 & 36.29 & 31.09 \\
            CoOp & \textbf{40.44} & 22.30 & 28.75 \\
            CoCoOp & 33.41 & 23.71 & 27.74 \\
            MaPLe & 37.44 & 35.61 & 36.50 \\
            \hline
            MetaPrompt & 40.14 & \textbf{36.51} & \textbf{38.24} \\
            vs. MaPLe & \textcolor{red}{+2.70} & \textcolor{red}{+0.90} & \textcolor{red}{+1.74} \\
            \hline
            \end{tabular}
            \vspace{\y pt}
	\end{minipage}\quad
 	\begin{minipage}{0.32\textwidth}
		\centering
            \normalsize
            (i) SUN397.
            \vspace{\x pt} 
            
		\begin{tabular}{lcc|c}
            \hline & \textbf{Base} & \textbf{New} & \textbf{H} \\
            \hline CLIP & 69.36 & 75.35 & 72.23 \\
            CoOp & 80.60 & 65.89 & 72.51 \\
            CoCoOp & 79.74 & 76.86 & 78.27 \\
            MaPLe & 80.82 & 78.70 & 79.75 \\
            \hline
            MetaPrompt & \textbf{82.26} & \textbf{79.04} & \textbf{80.62} \\
            vs. MaPLe & \textcolor{red}{+1.44} & \textcolor{red}{+1.34} & \textcolor{red}{+0.87} \\
            \hline
            \end{tabular}
            \vspace{\y pt}
	\end{minipage}

	\begin{minipage}{0.32\textwidth}
		\centering
            \normalsize
            (j) DTD.
            \vspace{\x pt} 
            
		\begin{tabular}{lcc|c}
            \hline & \textbf{Base} & \textbf{New} & \textbf{H} \\
            \hline CLIP &  53.24 & \textbf{59.90} & 56.37 \\
            CoOp & 79.44 & 41.18 & 54.24 \\
            CoCoOp & 77.01 & 56.00 & 64.85 \\
            MaPLe & 80.36 & 59.18 & 68.16 \\
            \hline
            MetaPrompt & \textbf{83.10} & 58.05 & \textbf{68.35} \\
            vs. MaPLe & \textcolor{red}{+2.74} & \textcolor{blue}{-1.13} & \textcolor{red}{+0.19} \\
            \hline
            \end{tabular}
            \vspace{\y pt}
	\end{minipage}\quad
        \begin{minipage}{0.32\textwidth}
		\centering
            \normalsize
            (k) EuroSAT.
            \vspace{\x pt} 
            
		\begin{tabular}{lcc|c}
            \hline & \textbf{Base} & \textbf{New} & \textbf{H} \\
            \hline CLIP & 56.48 & 64.05 & 60.03 \\
            CoOp & 92.19 & 54.74 & 68.69 \\
            CoCoOp & 87.49 & 60.04 & 71.21 \\
            MaPLe & \textbf{94.07} & 73.23 & 82.35 \\
            \hline
            MetaPrompt & 93.53 & \textbf{75.21} & \textbf{83.38} \\
            vs. MaPLe & \textcolor{blue}{-0.54} & \textcolor{red}{+1.98} & \textcolor{red}{+1.03} \\
            \hline
            \end{tabular}
            \vspace{\y pt}
	\end{minipage}\quad
 	\begin{minipage}{0.32\textwidth}
		\centering
            \normalsize
            (l) UCF101.
            \vspace{\x pt} 
            
		\begin{tabular}{lcc|c}
            \hline & \textbf{Base} & \textbf{New} & \textbf{H} \\
            \hline CLIP & 70.53 & 77.50 & 73.85 \\
            CoOp & 84.69 & 56.05 & 67.46 \\
            CoCoOp & 82.33 & 73.45 & 77.64 \\
            MaPLe & 83.00 & \textbf{78.66} & 80.77 \\
            \hline
            MetaPrompt & \textbf{85.33} & 77.72 & \textbf{81.35} \\
            vs. MaPLe & \textcolor{red}{+2.33} & \textcolor{blue}{-0.94} & \textcolor{red}{+0.58} \\
            \hline
            \end{tabular}
            \vspace{\y pt}
	\end{minipage}
\vspace{-5pt}
\end{table*}

\begin{table*}[t]
\caption{\textbf{Comparison of domain generalization methods and our MetaPrompt on four domain generalization benchmarks.} CLIP (template) indicates using ‘a photo of a \{class name\}’ prompt.
`Ensemble', `CLIP' indicate ensemble and CLIP-based methods. Our experiments are repeated three times using different random seeds. Although our method is based on \emph{few-shot} setting, it achieves competitive results against full-training methods and demonstrates strong performance on domain generalization benchmarks.\label{tab:dg}}
\centering
\small
\begin{tabular}{c|cc|cc|cccc}
    \hline 
    \multirow{2}*{\textbf{Method}} & \multicolumn{2}{c|}{\textbf{Setting}} & \multicolumn{2}{c|}{\textbf{Category}} & \multicolumn{4}{c}{\textbf{Accuracy(\%)}} \\
    \cline{2-9}
    ~ & \textbf{Zero-shot} & \textbf{Few-shot} & \textbf{Ensemble} & \textbf{CLIP} & \textbf{PACS} & \textbf{VLCS} & \textbf{OfficeHome} & \textbf{DomainNet}\\
    \hline 
    \multicolumn{1}{c|}{ERM \cite{gulrajani2020search}} & & \multicolumn{1}{c|}{ } & & \multicolumn{1}{c|}{ } & 84.2 $\pm$ 0.1 & 77.3 $\pm$ 0.1 & 67.6 $\pm$ 0.2 & 44.0 $\pm$ 0.1 \\
    \multicolumn{1}{c|}{MLDG \cite{li2018learning}} & & \multicolumn{1}{c|}{ } & & \multicolumn{1}{c|}{ } & 84.8 $\pm$ 0.6 & 77.1 $\pm$ 0.4 & 68.2 $\pm$ 0.1 & 41.8 $\pm$ 0.4 \\
    \multicolumn{1}{c|}{Fish \cite{shi2021gradient}} & & \multicolumn{1}{c|}{ } & & \multicolumn{1}{c|}{ } & 85.5 $\pm$ 0.3 & 77.8 $\pm$ 0.3 & 68.6 $\pm$ 0.4 & 42.7 $\pm$ 0.2 \\
    \multicolumn{1}{c|}{CORAL \cite{sun2016deep}} & & \multicolumn{1}{c|}{ } & & \multicolumn{1}{c|}{ } & 86.2 $\pm$ 0.3 & 78.8 $\pm$ 0.6 & 68.7 $\pm$ 0.3 & 41.5 $\pm$ 0.1 \\
    \multicolumn{1}{c|}{SWAD \cite{cha2021swad}} & & \multicolumn{1}{c|}{ } & $\checkmark$ & \multicolumn{1}{c|}{ } & 88.1 $\pm$ 0.1 & 79.1 $\pm$ 0.1 & 70.6 $\pm$ 0.2 & 46.5 $\pm$ 0.1 \\
    \multicolumn{1}{c|}{EoA \cite{arpit2021ensemble}} & & \multicolumn{1}{c|}{ } & $\checkmark$ & \multicolumn{1}{c|}{ } & 95.8 $\pm$ 0.0 & 81.1 $\pm$ 0.0 & 83.9 $\pm$ 0.0 & 60.9 $\pm$ 0.0 \\
    \multicolumn{1}{c|}{SEDGE \cite{li2022domain}} & & \multicolumn{1}{c|}{ } & $\checkmark$ & \multicolumn{1}{c|}{ } & 96.1 $\pm$ 0.0 & 82.2 $\pm$ 0.0 & 80.7 $\pm$ 0.2 & 54.7 $\pm$ 0.1 \\
    \multicolumn{1}{c|}{CLIP \cite{radford2021learning}} & $\checkmark$ & \multicolumn{1}{c|}{ }& & \multicolumn{1}{c|}{$\checkmark$} & 95.7 $\pm$ 0.0 & 75.9 $\pm$ 0.0 & 79.4 $\pm$ 0.0 & 57.9 $\pm$ 0.0 \\
    \multicolumn{1}{c|}{CLIP (template)} & $\checkmark$ & \multicolumn{1}{c|}{ } & & \multicolumn{1}{c|}{$\checkmark$} & 96.1 $\pm$ 0.0 & \textbf{82.3 $\pm$ 0.0} & 82.1 $\pm$ 0.0 & 57.8 $\pm$ 0.0 \\
    \multicolumn{1}{c|}{CoCoOp \cite{zhou2022conditional} (5-shot)} & & \multicolumn{1}{c|}{$\checkmark$} & & \multicolumn{1}{c|}{$\checkmark$} & 96.7 $\pm$ 0.4 & 78.3 $\pm$ 1.0 & 84.1 $\pm$ 0.1 & 61.1 $\pm$ 0.2 \\
    \hline
    \multicolumn{1}{c|}{MetaPrompt (1-shot)} & & \multicolumn{1}{c|}{$\checkmark$} & & \multicolumn{1}{c|}{$\checkmark$} & 96.7 $\pm$ 0.6 & 81.7 $\pm$ 0.6 & 84.0 $\pm$ 0.5 & 61.5 $\pm$ 0.2 \\
    \multicolumn{1}{c|}{MetaPrompt (5-shot)} & & \multicolumn{1}{c|}{$\checkmark$} & & \multicolumn{1}{c|}{$\checkmark$} & \textbf{96.9 $\pm$ 0.3} & 82.0 $\pm$ 0.9 & \textbf{85.1 $\pm$ 0.4} & \textbf{61.8 $\pm$ 0.2} \\
    \hline
\end{tabular}
\vspace{9pt}
\end{table*}

\subsection{Base-to-New Generalization}
The performance of our MetaPrompt in base-to-new generalization setting on 11 image recognition datasets is shown in Table \ref{tab:b2n}. 
We compare its performance with zero-shot CLIP with hand-crafted prompts and recent prompt learning methods, including CoOp, CoCoOp and MaPLe.
\paragraph{Generalization to Unseen Classes} 
In comparison with the state-of-the-art prompt tuning method MaPLe, MetaPrompt obtains an overall improvement to 75.48\% in terms of the average accuracy of new classes over 11 datasets with our episodic training strategy that explicitly constrains the prompt to generalize to out-of-domain classes.
When considering both base and new classes, MetaPrompt shows an absolute average gain of 0.54\% on the harmonic mean over MaPLe. The results strongly prove that our method of learning domain invariant prompt improves the generalization ability.
\paragraph{Performance Gain in Seen Classes} 
While MetaPrompt achieves excellent performance on generalizing to unseen classes, it still maintains high accuracy on seen classes compared with other methods optimized to fit in-domain data, even better than MaPLe by 1.37\%. 
MetaPrompt achieves a good trade-off between in-domain and out-of-domain data for two reasons. 
Firstly, our dual-modality prompts improve the recognition accuracy from two modalities simultaneously. With the unprompted pre-trained vision-language model as supervision, we obtain a stable boost in fitting both in-domain and out-of-domain data.
Secondly, from the perspective of training strategies, MaPLe does not explicitly consider the in-domain and out-domain trade-off and achieving good generalization at the expense of lower in-domain accuracy, while our approach proposes an explicit episodic training strategy to learn domain invariant prompt for both seen and unseen classes.

\subsection{Conventional Domain Generalization}
The performance of our MetaPrompt in conventional domain generalization setting on four benchmarks is shown in Table \ref{tab:dg}. We compare its performance with different categories of domain generalization methods, including the non-ensemble methods like ERM \cite{gulrajani2020search}, MLDG \cite{li2018learning}, Fish \cite{shi2021gradient}, CORAL \cite{sun2016deep}, the ensemble methods like SWAD \cite{cha2021swad}, EoA \cite{arpit2021ensemble}, SEDGE \cite{li2022domain}, as well as zero-shot CLIP and CoCoOp in domain generalization setting. Since extracting domain invariant features is the mainstream idea in traditional domain generalization tasks, we follow this idea for CLIP-based learning to train domain invariant prompt. 

In comparison with traditional domain generalization methods, CLIP-based methods show excellent generalization performance due to the strong transfer learning ability of CLIP.
Although training with very few samples, our MetaPrompt provides competitive results in domain generalization benchmarks, by outperforming all other methods on three of four benchmarks on average accuracy in the 5-shot setting and achieving comparable performance even with the 1-shot setting. 
Furthermore, our method outperforms the conditional prompt tuning method CoCoOp on all datasets, showing a much better generalization ability to unseen domains.
By simulating the generalization error between different domains during training, our domain invariant prompt is more generalizable than a conditional-based prompt generator training separately with domains.

But it is worth noting that our approach suffers some performance degradation in the few-shot regime on datasets with a large domain distribution shift, such as VLCS, which indicates that the current strategy still conducts an approximated estimate of generalization error.  

\begin{table}[t]
\caption{Ablation on different components. `Episodic' denotes our batch-wise episodic training strategy. `MOS' indicates using our modality-specific optimization strategy instead of regularizing prompts for both modalities in both tasks during episodic updates. For domain generalization, we use an average of 1-shot and 5-shot accuracy as evaluation metrics, the same below.\label{tab:md}}
\small
\centering
\center
(a) Base-to-New Generalization.
\vspace{5pt} 
    
\begin{tabular}{ccc|cc|c}
    \hline 
    \textbf{Episodic} & \textbf{AC Loss} & \textbf{MOS} & \textbf{Base} & \textbf{New} & \textbf{H}\\
    \hline 
     &  &  & 82.73 & 73.05 & 77.24 \\
    $\checkmark$ & &  & 82.88 & 74.96 & 78.62 \\
     & $\checkmark$ &  & 83.19 & 74.87 & 78.68 \\
    $\checkmark$ & $\checkmark$ &  & 83.60 & 75.22 & 78.91 \\
    $\checkmark$ & $\checkmark$ & $\checkmark$ & \textbf{83.65} & \textbf{75.48} & \textbf{79.09} \\
    \hline
\end{tabular}

\vspace{10pt}

\centering
\center
(b) Domain Generalization.
\vspace{5pt} 

\begin{tabular}{ccc|cccc}
    \hline 
    \textbf{Episodic} & \textbf{AC Loss} & \textbf{MOS} & \textbf{P} & \textbf{V} & \textbf{O} & \textbf{D}\\
    \hline 
     &  &  & 96.5 & 76.7 & 83.7 & 61.2 \\
     $\checkmark$ &  &  & 96.7 & 79.5 & 84.2 & 61.3\\
       & $\checkmark$ &  & 96.6 & 78.7 & 84.1 & 61.4 \\
     $\checkmark$ & $\checkmark$ &  & 96.7 & 81.0 & 84.4 & \textbf{61.6} \\
     $\checkmark$ & $\checkmark$ & $\checkmark$ & \textbf{96.8} & \textbf{81.8} & \textbf{84.5} & \textbf{61.6} \\
    \hline
\end{tabular}
\end{table}

\begin{figure*}[!t]
    \centering
    \center{\includegraphics[width=14cm]{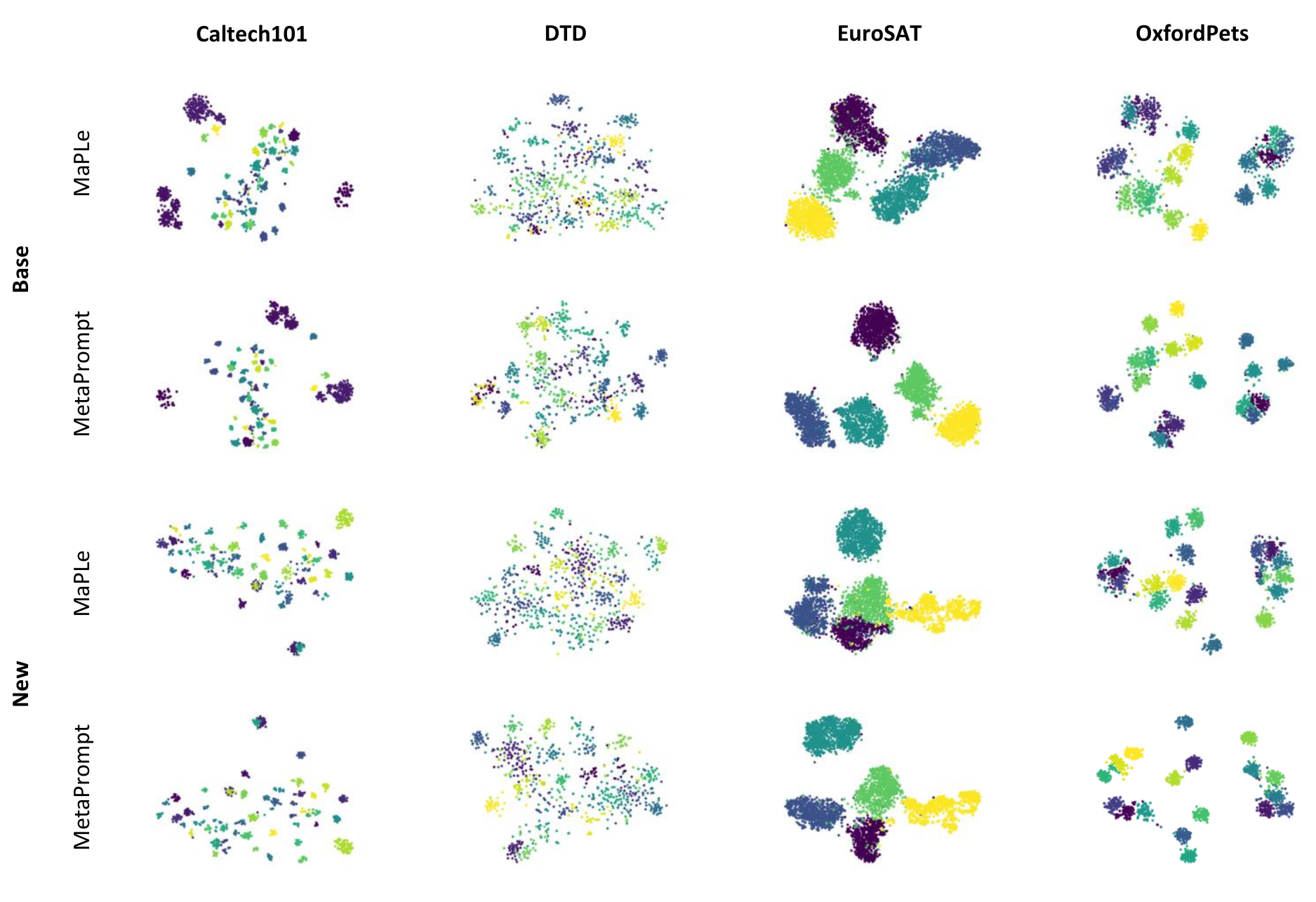}}
    \caption{T-SNE plots of image embeddings in previous methods MaPLe and our method MetaPrompt on four diverse image recognition datasets. Points with the same color represent image embeddings of the same class. MetaPrompt shows better inter-class separability and intra-class cohesiveness in both base and new classes.\label{fig:tsne}}
\end{figure*} 

\subsection{Further Analysis}
\paragraph{Influence of Model Components}
We analyze the influence of components in our model and conduct an ablation study on various combinations of them, as shown in Table \ref{tab:md}. The baseline method simultaneously trains both textual and visual prompts with a conventional gradient descent optimizer. The results show that both batch-wise episodic training strategy and asymmetric contrastive loss positively affect generalization to unseen domains. Among them, AC loss achieves an absolute performance gain on new class domains and an overall boost on new image domains, which shows the effectiveness of leveraging unprompted representations of pre-trained vision-language foundation models. Our episodic training strategy also plays an important role in boosting the ability of generalization, which will be analyzed in the subsequent section. In addition, our modality-specific optimization strategy further improves performance on both tasks.  

\paragraph{Visualization of Image Embeddings}
We randomly select four datasets to analyze the t-SNE plots of image embedding, as shown in Fig. \ref{fig:tsne}. Our MetaPrompt shows better inter-class separability and intra-class cohesiveness in both base and new classes. We attribute the good performance of our method to the fact that the visual prompts are learned under the supervision of textual representations of the unprompted pre-trained model. Since the representations remain fixed throughout the tuning process, visual concepts can be better aligned with corresponding textual labels. With the same class name, image embeddings are usually clustered together. On the other hand, the pre-trained CLIP model has a strong capability of representing semantics. With the supervision of distinguished textual semantics, image embeddings with various classes can be separated. 

\begin{table}[t]
\caption{Ablation on different prompt lengths.\label{tab:plength}}
\small
\centering
\center
(a) Base-to-New Generalization.
\vspace{5pt} 
    
\begin{tabular}{c|cc|c}
    \hline 
    \textbf{Length} & \textbf{Base} & \textbf{New} & \textbf{H}\\
    \hline 
    1 & 82.96 & 74.88 & 78.44 \\
    2 & 83.65 & \textbf{75.48} & \textbf{79.09} \\
    4 & 83.81 & 75.18 & 79.01 \\
    8 & 84.17 & 75.04 & 79.05 \\ 
    16 & 84.10 & 74.98 & 79.03 \\
    32 & \textbf{84.35} & 74.42 & 78.77 \\
    \hline
\end{tabular}

\vspace{10pt}

\center
(b) Domain Generalization.
\vspace{5pt} 

\begin{tabular}{c|cccc|c}
    \hline 
    \textbf{Length} & \textbf{P} & \textbf{V} & \textbf{O} & \textbf{D} & \textbf{Average}\\
    \hline 
    1 & 96.61 & 79.91 & 84.38 & 61.45 & 80.59 \\
    2 & 96.69 & 80.28 & 84.52 & 61.53 & 80.76 \\
    4 & 96.80 & \textbf{81.84} & \textbf{84.54} & 61.63 & \textbf{81.20} \\
    8 & 96.92 & 80.49 & 84.35 & 61.50 & 80.82 \\
    16 & \textbf{96.93} & 79.75 & 84.49 & 61.60 & 80.69 \\
    32 & 96.87 & 79.28 & 84.48 & \textbf{61.67} & 80.57 \\
    \hline
\end{tabular}
\end{table}

\begin{table}[t]
\caption{Ablation on different layers of prompt.\label{tab:player}}
\small
\centering
\center
(a) Base-to-New Generalization.
\vspace{5pt} 
    
\begin{tabular}{c|cc|c}
    \hline 
    \textbf{Layer} & \textbf{Base} & \textbf{New} & \textbf{H}\\
    \hline 
    2 & 80.32 & 74.87 & 77.19 \\
    4 & 81.09 & 74.27 & 77.21 \\
    6 & 81.89 & \textbf{75.53} & 78.35 \\
    8 & 82.57 & 75.01 & 78.32 \\
    10 & 83.22 & 75.51 & 78.90 \\ 
    12 & \textbf{83.65} & 75.48 & \textbf{79.09} \\
    \hline
\end{tabular}

\vspace{10pt}

\center
(b) Domain Generalization.
\vspace{5pt} 

\begin{tabular}{c|cccc|c}
    \hline 
    \textbf{Layer} & \textbf{P} & \textbf{V} & \textbf{O} & \textbf{D} & \textbf{Average}\\
    \hline 
    2 & 96.66 & 81.32 & 83.96 & 60.67 & 80.65 \\
    4 & 96.65 & 81.51 & 84.11 & 60.93 & 80.80 \\
    6 & 96.59 & 81.31 & 84.30 & 61.10 & 80.82 \\
    8 & 96.47 & 80.54 & 84.62 & 61.31 & 80.73 \\
    10 & \textbf{96.80} & \textbf{81.84} & 84.54 & \textbf{61.63} & \textbf{81.20} \\
    12 & 96.64 & 78.35 & \textbf{84.85} & 61.51 & 80.34 \\
    \hline
\end{tabular}
\end{table}

\paragraph{Influence of Prompt Length}
The ablation study on the prompt length is carried out in both generalization settings. We study 1, 2, 4, 8, 16, and 32 prompt vectors each layer for both modalities with the same random initialization. 
Table \ref{tab:plength} summarizes the performance over both tasks.
For base-to-new generalization, we can draw a conclusion that the model with a longer prompt length performs better in base classes. 
On the other hand, by applying our training strategy, the difference in new classes is relatively small except for 32 prompt vectors with a dramatic drop in performance.
The result suggests that using 2 prompt vectors is a better choice when taking into account the accuracy of both base and new classes.
For conventional domain generalization, a shorter prompt is not enough to recognize visual concepts well, while a longer prompt seems to overfit in-domain samples. With a prompt length of 4, our method shows promising results considering the overall performance.

\paragraph{Influence of Layer of Prompt }
The ablation study on the layer of prompt is also conducted in both generalization settings. We study 2, 4, 6, 8, 10, and 12 layers of prompts for both modalities with the same random initialization. 
Table \ref{tab:player} demonstrates the results for these settings.
For base-to-new generalization, we can draw a similar conclusion that the accuracy of base classes improves as the layer of prompts increases, while the results on new classes show instability when it changes.
The result indicates that applying prompts to all 12 layers has a strong performance when taking into account both base and new classes.
For conventional domain generalization, it is clear that with 10 layers of prompt, our method shows excellent performance on all datasets.

\begin{figure}[t]
  \centering
    \center{\includegraphics[width=8cm]
    {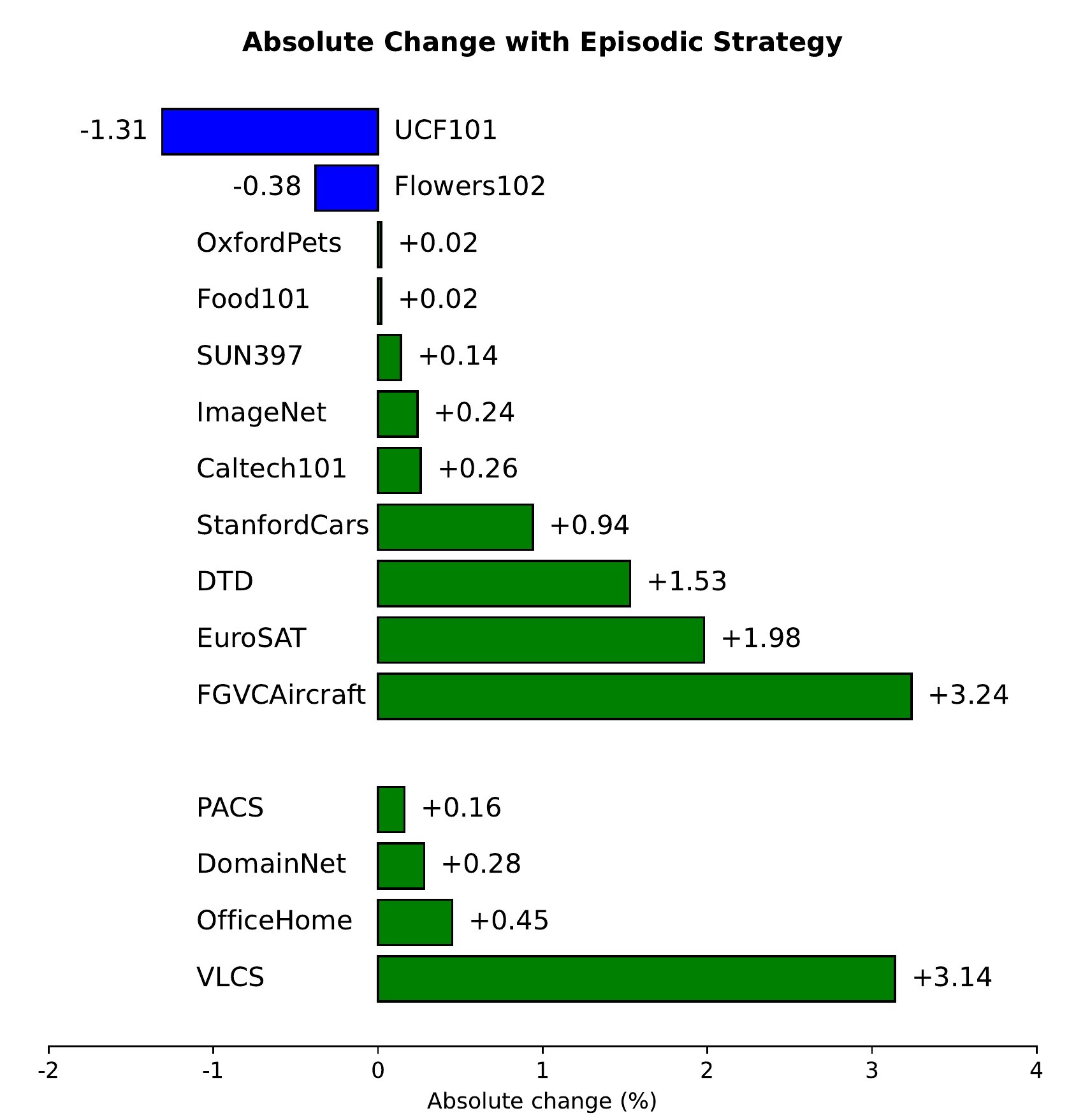}}
    \vspace{-5pt}
    \caption{Performance change on unseen domains with our proposed episodic training over datasets for base-to-new generalization and domain generalization. It shows a general improvement in both generalization tasks, which proves the effectiveness of our episodic training. \label{fig:epi}}
\end{figure}

\paragraph{Influence of Episodic Training}
We investigate the influence of our proposed batch-wise episodic training strategy. Fig. \ref{fig:epi} demonstrates an overall performance boost on datasets for both generalization tasks. The performance of our training strategy remains robust on new classes, which reflects excellent generalization ability. For FGVCAircraft in base-to-new generalization and VLCS in conventional generalization, our training strategy improves the accuracy by more than 3\%, which prevents catastrophic failures on out-domain data, highlighting the significance of learning domain invariant prompt. 

\section{Conclusion}
We propose MetaPrompt for learning domain invariant prompt with CLIP to tackle the problem of generalization. Our theoretical analysis shows that meta-learning applying the episodic training strategy has a strong generalization guarantee. Based on this analysis, we design a dual-modality prompt tuning network with asymmetric contrastive loss and impose a batch-wise episodic training strategy as an explicit constraint on prompt tuning. 
Prompt can be learned on few-shot data with a high generalization ability to unseen classes and domains. Extensive experiments on base-to-new generalization and domain generalization demonstrate that our method consistently outperforms existing methods. 

While traditional prompt learning approaches often degrade the performance on generalization, our method provides timely insights on how to access the intrinsic association between domains and proposes a feasible solution for learning invariant prompts, which alleviates poor performance on unseen tasks. We reveal that MetaPrompt performs much better in many generalization tasks than input-conditional approaches and show evidence that learning domain invariant prompts has the potential for large pre-trained foundation models. We hope the empirical findings could inspire future research on invariant prompt learning for efficient generalization.   

{\small
\bibliographystyle{ieee_fullname}
\bibliography{egbib}
}

\begin{IEEEbiography}
[{\includegraphics[width=1in,height=1.25in,clip,keepaspectratio]{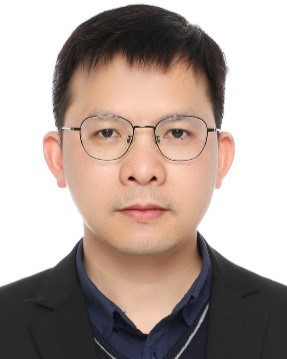}}] 
{Cairong Zhao} is currently a Professor of College of Electronic and Information Engineering at Tongji University. He received a Ph.D. degree from Nanjing University of Science and Technology, an M.S. degree from Changchun Institute of Optics, Fine Mechanics and Physics, Chinese Academy of Sciences and a B.S. degree from Jilin University, in 2011, 2006 and 2003, respectively. He works on visual and intelligent learning, including computer vision, pattern recognition and visual surveillance. He has published over 40 top-rank international conferences and journals in the field, including CVPR, ICCV, ICLR, AAAI, ACM MM, TIP, TIFS, TMM, TCSVT, and PR. He holds prestigious positions such as the deputy secretary-general of the Pattern Recognition and Machine Intelligence Committee of the Chinese Association of Automation, the chairman of the Computer Vision Special Committee of the Shanghai Computer Society, and an outstanding member of the China Computer Federation, and a senior member of the China Graphics Society. He also serves as the reviewer of more than ten AI-related international journals and conferences, including TPAMI, TIP, CVPR, ICCV, NIPS, ICML, AAAI, etc.
\end{IEEEbiography}

\begin{IEEEbiography}
[{\includegraphics[width=1in,height=1.25in,clip,keepaspectratio]{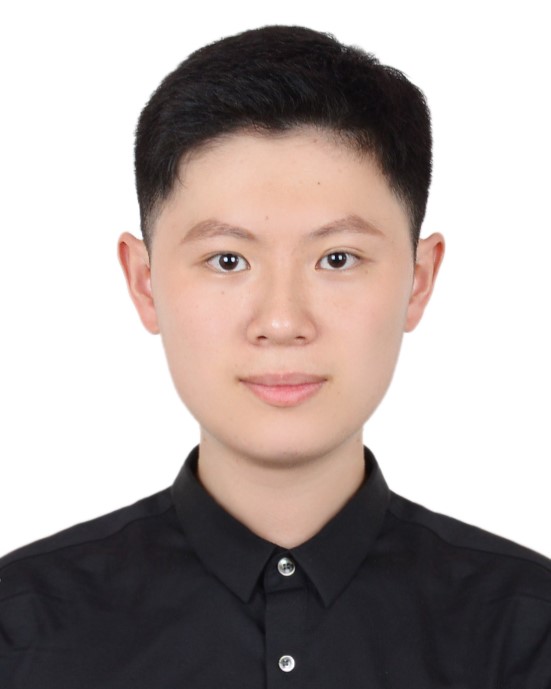}}] 
{Yubin Wang} received the B.E. degree in data science and big data technology from Tongji University in 2022. He is currently pursuing the master’s degree with Tongji University. His main research interests include prompt learning, multi-modal learning, and person re-identification.
\end{IEEEbiography}

\begin{IEEEbiography}
[{\includegraphics[width=1in,height=1.25in,clip,keepaspectratio]{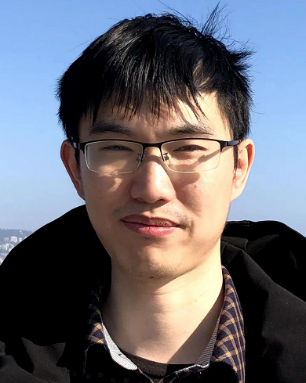}}] 
{Xinyang Jiang} received B.E. from Zhejiang University in 2012 and Ph.D. from Zhejiang University in 2017. He is now a researcher from Microsoft Research Asia. Before joining MSRA, he was a researcher from Tencent Youtu Lab. His main research field is computer vision, including person Re-identification, vector graphics recognition and video enhancement and recognition.
\end{IEEEbiography}

\begin{IEEEbiography}
[{\includegraphics[width=1in,height=1.25in,clip,keepaspectratio]{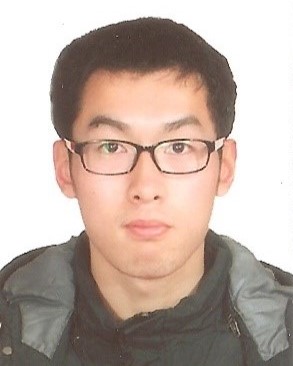}}] 
{Yifei Shen} (Graduate Student Member, IEEE) received the Ph.D. degree from the Hong Kong University of Science and Technology and currently works at Microsoft.
\end{IEEEbiography}

\begin{IEEEbiography}
[{\includegraphics[width=1in,height=1.25in,clip,keepaspectratio]{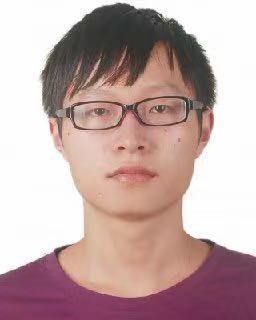}}] 
{Kaitao Song} received the B.S. degree and Ph.D. degree in computer science and technology from Nanjing University of Science and Technology, China, in 2015 and 2021. His current research interests include natural language processing, multimodal analysis, deep learning, speech recognition and machine learning. He has published more than 20 academic papers on the top-tier international journals and conferences, such as IEEE TIP, ICML, NeurIPS, ACL, KDD, ICCV, AAAI, IJCAI, InterSpeech, ICASSP, etc. He has served as a PC member of ICML, NeurIPS, ICLR, ACL, EMNLP and etc.
\end{IEEEbiography}

\begin{IEEEbiography}
[{\includegraphics[width=1in,height=1.25in,clip,keepaspectratio]{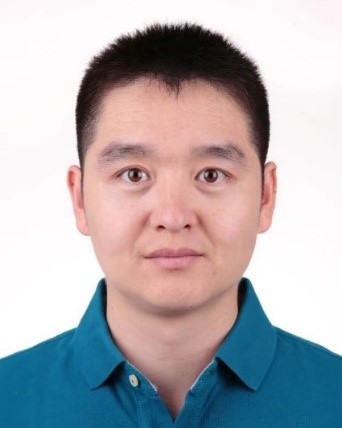}}] 
{Dongsheng Li} received B.E. from University of Science and Technology of China in 2007 and Ph.D. from Fudan University in 2012. He is now a principal research manager with Microsoft Research Asia (MSRA) since February 2020.  Before joining MSRA, he was a research staff member with IBM Research – China since April 2015. He is also an adjunct professor with School of Computer Science, Fudan University, Shanghai, China. His research interests include recommender systems and machine learning applications. His work on cognitive recommendation engine won the 2018 IBM Corporate Award.
\end{IEEEbiography}

\begin{IEEEbiography}
[{\includegraphics[width=1in,height=1.25in,clip,keepaspectratio]{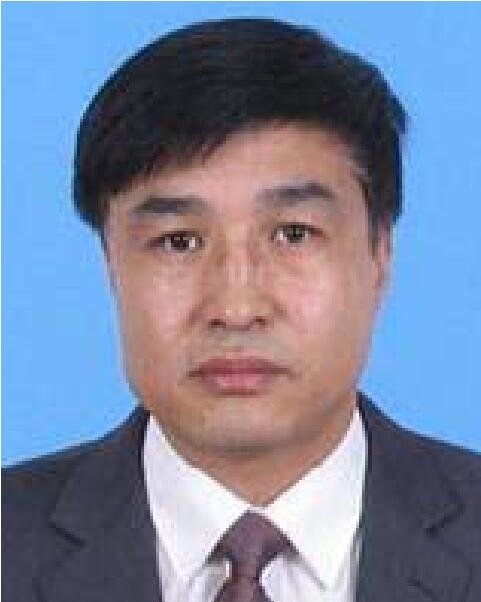}}] 
{Duoqian Miao}  was born in 1964. He is currently a Professor and the Ph.D. Tutor with the College of Electronics and Information Engineering, Tongji University, Shanghai, China. He serves as the Vice President for the International Rough Set Society, the Executive Manager of the Chinese Association for Artificial Intelligence, the Chair of the CAAI Granular Computing Knowledge Discovery Technical Committee, a Distinguished Member of Chinese Computer Federation, the Vice President of the Shanghai Computer Federation, and the Vice President of the Shanghai Association for Artificial Intelligence. He serves as Associate Editor for the International Journal of Approximate Reasoning and an Editor of the Journal of Computer Research and Development (in Chinese).
\end{IEEEbiography}

\end{document}